\documentclass{article}

\usepackage{PRIMEarxiv}

\usepackage[utf8]{inputenc} % allow utf-8 input
\usepackage[T1]{fontenc}    % use 8-bit T1 fonts
\usepackage{hyperref}       % hyperlinks
\usepackage{url}            % simple URL typesetting
\usepackage{booktabs}       % professional-quality tables
\usepackage{amsfonts}       % blackboard math symbols
\usepackage{nicefrac}       % compact symbols for 1/2, etc.
\usepackage{microtype}      % microtypography
\usepackage{lipsum}
\usepackage{fancyhdr}       % header
\usepackage{graphicx}       % graphics
\graphicspath{{media/}}     % organize your images and other figures under media/ folder
\usepackage{amsmath}
\usepackage{listings}

\usepackage{algorithm}
\usepackage{algorithmic}

\title{End-to-end Sequence Labeling via Bi-directional LSTM-CNNs-CRF: \\
A Reproducibility Study
}

\author{
  Anirudh Ganesh \\
  The Ohio State University \\
  Columbus, OH\\
  \texttt{ganesh.48@osu.edu} \\
   \And
  Jayavardhan Reddy \\
  The Ohio State University \\
  Columbus, OH\\
  \texttt{peddamail.1@osu.edu} \\
}

\begin{document}
\maketitle

\begin{abstract}
We present a reproducibility study of the state-of-the-art neural architecture for sequence labeling proposed by Ma and Hovy (2016)\cite{ma2016end}. The original BiLSTM-CNN-CRF model combines character-level representations via Convolutional Neural Networks (CNNs), word-level context modeling through Bi-directional Long Short-Term Memory networks (BiLSTMs), and structured prediction using Conditional Random Fields (CRFs). This end-to-end approach eliminates the need for hand-crafted features while achieving excellent performance on named entity recognition (NER) and part-of-speech (POS) tagging tasks. Our implementation successfully reproduces the key results, achieving 91.18\% F1-score on CoNLL-2003 NER and demonstrating the model's effectiveness across sequence labeling tasks. We provide a detailed analysis of the architecture components and release an open-source PyTorch implementation to facilitate further research.
\end{abstract}

% keywords can be removed

\section{Introduction}
Sequence labeling is a fundamental task in natural language processing that involves assigning labels to each token in a sequence. Traditional approaches to sequence labeling tasks such as named entity recognition (NER) and part-of-speech (POS) tagging rely heavily on hand-crafted features and domain-specific preprocessing pipelines. These methods are often labor-intensive, require expert knowledge, and may not generalize well across different domains or languages.

The seminal work by Ma and Hovy (2016) introduced an end-to-end neural architecture that addresses these limitations by combining three key components. The architecture uses character-level representations learned through Convolutional Neural Networks (CNNs) to capture morphological information, employs word-level context modeling with Bi-directional Long Short-Term Memory networks (BiLSTMs), and applies structured prediction via Conditional Random Fields (CRFs) to ensure coherent label sequences.

This architecture has significantly impacted the NLP community, inspiring numerous follow-up works and establishing itself as a standard baseline for sequence labeling tasks. The model achieved state-of-the-art results on multiple benchmarks, attaining 91.21\% F1-score on the CoNLL-2003 NER dataset and 97.55\% accuracy on Penn Treebank WSJ POS tagging.

In this paper, we present a comprehensive reproducibility study of the BiLSTM-CNN-CRF architecture. Our work provides a detailed analysis of each architectural component and its role in the overall system. We present a complete PyTorch implementation that reproduces the original results and conduct extensive experimental validation on standard benchmarks. We also discuss implementation details and hyperparameter sensitivity while releasing our open-source code to facilitate further research.

\section{Related Work}

Traditional sequence labeling approaches typically employ feature-based methods combined with probabilistic models such as Hidden Markov Models (HMMs) or Conditional Random Fields \cite{lafferty2001conditional}. These methods require extensive feature engineering, including orthographic features, dictionaries, and hand-crafted rules.

The introduction of neural approaches began with feedforward networks\cite{collobert2011natural} and evolved to include recurrent architectures. Huang et al. (2015)\cite{huang2015bidirectional} first demonstrated the effectiveness of BiLSTM-CRF models for sequence labeling, while Santos and Zadrozny (2014)\cite{santos2014learning} showed the utility of character-level CNNs for capturing morphological information.

The BiLSTM-CNN-CRF architecture builds upon these foundations by integrating character and word-level representations in an end-to-end framework. Subsequent work has extended this architecture with attention mechanisms (Rei et al., 2016)\cite{rei2016attending}, contextualized embeddings (Peters et al., 2018)\cite{peters2018deep}, and transformer-based models (Devlin et al., 2019)\cite{devlin2019bert}.

\section{Model Architecture}

The BiLSTM-CNN-CRF model consists of three main components that operate in sequence. These components include character-level CNN encoding, word-level BiLSTM encoding, and CRF-based structured prediction. 

\subsection{Character-level CNN Representation}

For each word, character-level features are extracted using a CNN to capture morphological patterns such as prefixes, suffixes, and capitalization. Given a word $w$ with characters $c_1, c_2, \ldots, c_m$, each character is first embedded into a $d_c$-dimensional vector space using a character embedding matrix $E^c \in \mathbb{R}^{|C| \times d_c}$, where $|C|$ is the character vocabulary size.

The character embeddings are then fed into a 1D convolutional layer with kernel size $k$ and $n_f$ filters. For a window of characters $c_{i:i+k-1}$, the convolution operation produces:

\begin{equation}
h_i = \tanh(W^c \cdot c_{i:i+k-1} + b^c)
\end{equation}

where $W^c \in \mathbb{R}^{n_f \times (k \cdot d_c)}$ and $b^c \in \mathbb{R}^{n_f}$. Max-pooling is applied over the entire word to obtain the final character-level representation:

\begin{equation}
r^c(w) = \max_{1 \leq i \leq m-k+1} h_i
\end{equation}

\subsection{Word-level BiLSTM Encoding}

Each word is represented by concatenating its pre-trained word embedding with its character-level CNN representation:

\begin{equation}
x_t = [r^w(w_t); r^c(w_t)]
\end{equation}

where $r^w(w_t)$ is the word embedding for word $w_t$ and $[;]$ denotes concatenation.

The concatenated representations are fed into a bidirectional LSTM to capture contextual information from both directions:

\begin{equation}
\overrightarrow{h_t} = \text{LSTM}(x_t, \overrightarrow{h_{t-1}})
\end{equation}

The final hidden representation is obtained by concatenating the forward and backward states:

\begin{equation}
h_t = [\overrightarrow{h_t}; \overleftarrow{h_t}]
\end{equation}

These hidden states are then passed through a linear layer to produce tag scores:

\begin{equation}
s_t = W^s h_t + b^s
\end{equation}

where $s_t \in \mathbb{R}^{|T|}$ and $|T|$ is the number of possible tags.

\subsection{CRF Layer}

While the BiLSTM produces local tag scores, it doesn't consider dependencies between consecutive tags. The CRF layer addresses this by modeling the conditional probability of the entire tag sequence.

For a sentence $x = (x_1, x_2, \ldots, x_n)$ and corresponding tag sequence $y = (y_1, y_2, \ldots, y_n)$, the CRF defines a global score:

\begin{equation}
\text{Score}(x, y) = \sum_{i=1}^{n} s_i[y_i] + \sum_{i=1}^{n-1} T[y_i, y_{i+1}] + b[y_1] + e[y_n]
\end{equation}

where $s_i[y_i]$ is the emission score for tag $y_i$ at position $i$, $T[y_i, y_{i+1}]$ is the transition score from tag $y_i$ to $y_{i+1}$, and $b, e$ are begin and end tag scores.

The conditional probability is then:

\begin{equation}
P(y|x) = \frac{\exp(\text{Score}(x, y))}{\sum_{y'} \exp(\text{Score}(x, y'))}
\end{equation}

During training, we maximize the log-likelihood of the correct tag sequence. During inference, we use the Viterbi algorithm to find the most probable tag sequence.

\section{Experimental Setup}

\subsection{Datasets}

We evaluate our model on two standard sequence labeling benchmarks. The CoNLL-2003 NER English dataset contains four entity types (PER, LOC, ORG, MISC) with 14,987 training sentences, 3,466 development sentences, and 3,684 test sentences. The Penn Treebank WSJ POS dataset contains 45 POS tags with 39,832 training sentences, 1,700 development sentences, and 2,416 test sentences.

\subsection{Data Preprocessing}

Following the original paper, we apply various preprocessing steps. First, we convert the BIO tagging scheme to BIOES for finer granularity. We then replace all digits with '0' to reduce sparsity. Words are lowercased when controlled by the hyperparameter setting. Finally, character sequences are padded to the maximum word length for batch processing.

The BIOES scheme extends BIO by adding explicit end (E) and single (S) tags, providing better boundary information for multi-token entities.

\subsection{Model Configuration}

Our model configuration matches the original paper. The character embedding dimension is set to 30, while the word embedding dimension uses 100-dimensional GloVe 6B vectors. The character CNN employs 30 filters with a kernel size of 3. The BiLSTM hidden dimension is 200 (100 for each direction). We apply a dropout rate of 0.5 and use SGD optimization with a learning rate of 0.015 and momentum of 0.9. Gradient clipping is applied at 5.0, with a batch size of 10 and a maximum of 50 training epochs.

\subsection{Training Procedure}

We train the model using Stochastic Gradient Descent (SGD) with momentum. The negative log-likelihood serves as the loss function:

\begin{equation}
\mathcal{L} = -\log P(y^{(i)}|x^{(i)})
\end{equation}

Early stopping is applied based on development set F1-score, with a patience of 10 epochs. Gradient clipping prevents exploding gradients during training.

\section{Results}

\subsection{Named Entity Recognition}

Table~\ref{tab:ner_results} shows our results on CoNLL-2003 NER compared to the original paper and other baseline methods.

\begin{table}[h]
\centering
\begin{tabular}{lc}
\hline
\textbf{Method} & \textbf{F1} \\
\hline
CRF & 84.04 \\
BiLSTM & 88.83 \\
BiLSTM-CRF & 90.10 \\
CNN-BiLSTM-CRF & 90.94 \\
\hline
BiLSTM-CNN-CRF (original) & 91.21 \\
BiLSTM-CNN-CRF (ours) & 91.18 \\
\hline
\end{tabular}
\caption{F1-scores on CoNLL-2003 NER test set}
\label{tab:ner_results}
\end{table}

Our implementation achieves 91.18\% F1-score, which closely matches the original 91.21\% result. The slight difference can be attributed to random initialization and implementation details.

\subsection{Part-of-Speech Tagging}

On Penn Treebank WSJ POS tagging, our model achieves 97.52\% accuracy compared to the original 97.55\%, confirming successful reproduction.

\subsection{Ablation Study}

To understand the contribution of each component, we perform an ablation study:

\begin{table}[h]
\centering
\begin{tabular}{lc}
\hline
\textbf{Configuration} & \textbf{F1} \\
\hline
Word embeddings only & 85.23 \\
+ Character CNN & 89.67 \\
+ BiLSTM & 90.83 \\
+ CRF & 91.18 \\
\hline
\end{tabular}
\caption{Ablation study on CoNLL-2003 NER}
\label{tab:ablation}
\end{table}

Each component provides substantial improvements. The CRF layer contributes 0.35 F1 points by ensuring label consistency.

\section{Implementation Details}

\subsection{Technical Challenges}

Multiple implementation challenges arose during reproduction. The original paper does not specify batching details, leading us to implement dynamic batching with padding for efficient GPU utilization. Our CNN-based approach for character-level representations required careful padding and masking for variable-length words. The CRF layer implementation also demanded efficient computation of the partition function using the forward algorithm and Viterbi decoding for inference.

\subsection{PyTorch Implementation}

Our PyTorch implementation offers multiple advantages through its modular design that allows easy experimentation. The implementation includes GPU support for efficient training and provides comprehensive evaluation metrics. We also provide pretrained model checkpoints along with detailed documentation and examples.

The complete implementation is available at \url{https://github.com/TheAnig/NER-LSTM-CNN-Pytorch}.

\section{Discussion}

\subsection{Model Analysis}

The BiLSTM-CNN-CRF architecture demonstrates notable strengths. The character-level modeling through the CNN component effectively captures morphological patterns, proving particularly beneficial for handling out-of-vocabulary words and languages with rich morphology. The contextual encoding via the bidirectional LSTM captures long-range dependencies and contextual information crucial for disambiguation. The structured prediction capabilities of the CRF layer ensure globally coherent predictions, preventing impossible tag transitions such as I-PER following I-ORG.

\subsection{Limitations}

Despite its effectiveness, the model has notable limitations. The computational complexity increases with sequence length, and the model has limited ability to handle very long sequences due to LSTM constraints. The architecture requires careful hyperparameter tuning and may struggle with domain transfer without fine-tuning.

\subsection{Impact and Future Directions}

The BiLSTM-CNN-CRF architecture has significantly influenced subsequent research in sequence labeling. Modern approaches build upon this foundation by incorporating transformer-based models such as BERT and RoBERTa, utilizing contextualized embeddings like ELMo and FLAIR, implementing multi-task learning frameworks, and exploring cross-lingual transfer learning techniques.

\section{Reproducibility Considerations}

Our reproducibility study highlights important factors that affect result consistency. Minor differences in implementation details such as initialization, batching, and optimization can significantly affect results, which is why we provide comprehensive implementation details to aid reproduction. The model exhibits sensitivity to learning rate and dropout settings, and we include hyperparameter sweep results in our repository to document this behavior. Evaluation consistency also plays a crucial role, as different evaluation scripts can yield slightly different results; therefore, we use the standard CoNLL evaluation script for consistency. Finally, hardware dependencies between GPU and CPU training can introduce minor variations due to numerical precision differences.

\section{Conclusion}

We have successfully reproduced the BiLSTM-CNN-CRF architecture for sequence labeling, achieving results that closely match the original paper. Our implementation demonstrates the effectiveness of combining character-level CNNs, word-level BiLSTMs, and CRF structured prediction in an end-to-end framework.

The model's strong performance on NER and POS tagging tasks, combined with its conceptual simplicity, explains its widespread adoption in the NLP community. Our open-source implementation and detailed analysis facilitate further research and applications in sequence labeling.

Future work may explore integrating modern contextualized embeddings while maintaining the architectural principles that make this model effective. This reproducibility study underscores the importance of comprehensive implementation details and evaluation consistency in neural NLP research.

\section*{Acknowledgments}

We thank the original authors for their groundbreaking work and the open-source community for developing the tools that made this reproduction possible. We also acknowledge the computational resources provided by our institution.

\nocite{ratinov2009design}
\nocite{toutanova2003feature}
\nocite{tjong2003introduction}
\nocite{marcus1993building}
\nocite{pennington2014glove}

\bibliographystyle{plain}
\bibliography{references}

\newpage

\section*{Appendix}
To aid researchers in replicating the current state-of-the-art, this appendix contains comprehensive documentation and additional data, allowing for a more complete understanding of the experimental setup.

\subsection*{Data Statistics}

\subsubsection*{CoNLL-2003 NER Dataset}

\begin{table}[h]
\centering
\begin{tabular}{lrrr}
\toprule
\textbf{Split} & \textbf{Sentences} & \textbf{Tokens} & \textbf{Entities} \\
\midrule
Train & 14,987 & 203,621 & 23,499 \\
Dev & 3,466 & 51,362 & 5,942 \\
Test & 3,684 & 46,435 & 5,648 \\
\bottomrule
\end{tabular}
\caption{CoNLL-2003 dataset statistics}
\label{tab:conll_stats}
\end{table}

\begin{table}[h]
\centering
\begin{tabular}{lrrrr}
\toprule
\textbf{Entity Type} & \textbf{Train} & \textbf{Dev} & \textbf{Test} & \textbf{Total} \\
\midrule
PER & 6,600 & 1,842 & 1,617 & 10,059 \\
LOC & 7,140 & 1,837 & 1,668 & 10,645 \\
ORG & 6,321 & 1,341 & 1,661 & 9,323 \\
MISC & 3,438 & 922 & 702 & 5,062 \\
\bottomrule
\end{tabular}
\caption{Entity distribution in CoNLL-2003}
\label{tab:entity_dist}
\end{table}

\subsubsection*{Penn Treebank WSJ POS Dataset}

\begin{table}[h]
\centering
\begin{tabular}{lrr}
\toprule
\textbf{Split} & \textbf{Sentences} & \textbf{Tokens} \\
\midrule
Train & 39,832 & 950,028 \\
Dev & 1,700 & 40,117 \\
Test & 2,416 & 56,684 \\
\bottomrule
\end{tabular}
\caption{Penn Treebank WSJ dataset statistics}
\label{tab:ptb_stats}
\end{table}

\subsection*{Implementation Details}

\subsubsection*{Character CNN Implementation}

\begin{lstlisting}[language=Python, caption=Character CNN layer implementation]
class CharCNN(nn.Module):
    def __init__(self, char_vocab_size, char_emb_dim, 
                 char_hidden_dim, dropout=0.5):
        super(CharCNN, self).__init__()
        self.char_emb_dim = char_emb_dim
        self.char_hidden_dim = char_hidden_dim
        
        # Character embeddings
        self.char_embeds = nn.Embedding(char_vocab_size, char_emb_dim)
        
        # CNN layers with different kernel sizes
        self.char_cnn = nn.Conv2d(in_channels=1, 
                                  out_channels=char_hidden_dim,
                                  kernel_size=(3, char_emb_dim), 
                                  padding=(2, 0))
        
        self.dropout = nn.Dropout(dropout)
        
    def forward(self, chars):
        # chars: (batch_size, max_word_len)
        char_embeds = self.char_embeds(chars).unsqueeze(1)
        # char_embeds: (batch_size, 1, max_word_len, char_emb_dim)
        
        conv_out = self.char_cnn(char_embeds)
        # conv_out: (batch_size, char_hidden_dim, new_len, 1)
        
        # Max pooling over sequence length
        pooled = F.max_pool2d(conv_out, 
                              kernel_size=(conv_out.size(2), 1))
        # pooled: (batch_size, char_hidden_dim, 1, 1)
        
        char_repr = pooled.squeeze(-1).squeeze(-1)
        # char_repr: (batch_size, char_hidden_dim)
        
        return self.dropout(char_repr)
\end{lstlisting}

\subsubsection*{BiLSTM Implementation}

\begin{lstlisting}[language=Python, caption=BiLSTM layer implementation]
class BiLSTM(nn.Module):
    def __init__(self, input_dim, hidden_dim, num_layers=1, 
                 dropout=0.5):
        super(BiLSTM, self).__init__()
        self.hidden_dim = hidden_dim
        self.num_layers = num_layers
        
        self.lstm = nn.LSTM(input_dim, hidden_dim // 2,
                            num_layers=num_layers,
                            bidirectional=True,
                            dropout=dropout if num_layers > 1 else 0,
                            batch_first=True)
        
        self.dropout = nn.Dropout(dropout)
        
    def forward(self, embeddings, lengths):
        # Pack padded sequences
        packed = nn.utils.rnn.pack_padded_sequence(
            embeddings, lengths, batch_first=True, 
            enforce_sorted=False)
        
        # BiLSTM forward pass
        packed_output, (hidden, cell) = self.lstm(packed)
        
        # Unpack sequences
        output, _ = nn.utils.rnn.pad_packed_sequence(
            packed_output, batch_first=True)
        
        return self.dropout(output)
\end{lstlisting}

\subsubsection*{CRF Implementation}

\begin{lstlisting}[language=Python, caption=CRF layer implementation]
class CRF(nn.Module):
    def __init__(self, num_tags, batch_first=True):
        super(CRF, self).__init__()
        self.num_tags = num_tags
        self.batch_first = batch_first
        
        # Transition parameters
        self.transitions = nn.Parameter(torch.randn(num_tags, num_tags))
        
        # Start and end transitions
        self.start_transitions = nn.Parameter(torch.randn(num_tags))
        self.end_transitions = nn.Parameter(torch.randn(num_tags))
        
        self.reset_parameters()
        
    def reset_parameters(self):
        nn.init.uniform_(self.transitions, -0.1, 0.1)
        nn.init.uniform_(self.start_transitions, -0.1, 0.1)
        nn.init.uniform_(self.end_transitions, -0.1, 0.1)
        
    def forward(self, emissions, tags, mask=None):
        """Compute the conditional log likelihood of tag sequences"""
        if mask is None:
            mask = torch.ones_like(tags, dtype=torch.bool)
            
        if self.batch_first:
            emissions = emissions.transpose(0, 1)
            tags = tags.transpose(0, 1)
            mask = mask.transpose(0, 1)
            
        # Compute normalization constant (partition function)
        numerator = self._compute_score(emissions, tags, mask)
        denominator = self._compute_normalizer(emissions, mask)
        
        return torch.sum(numerator - denominator)
        
    def decode(self, emissions, mask=None):
        """Find the most likely tag sequence using Viterbi algorithm"""
        if mask is None:
            mask = torch.ones(emissions.shape[:2], dtype=torch.bool, 
                              device=emissions.device)
            
        if self.batch_first:
            emissions = emissions.transpose(0, 1)
            mask = mask.transpose(0, 1)
            
        return self._viterbi_decode(emissions, mask)
\end{lstlisting}
\newpage
\subsection*{Training Algorithm}

\begin{algorithm}
\caption{BiLSTM-CNN-CRF Training}
\label{alg:training}
\begin{algorithmic}
\REQUIRE Training data $D = \{(x^{(i)}, y^{(i)})\}_{i=1}^N$
\REQUIRE Hyperparameters: learning rate $\eta$, batch size $B$, epochs $E$
\STATE Initialize model parameters $\theta$
\STATE Load pre-trained word embeddings
\FOR{epoch = 1 to $E$}
    \STATE Shuffle training data $D$
    \FOR{each batch $B_j$ in $D$}
        \STATE $\mathcal{L} = 0$
        \FOR{each example $(x, y)$ in $B_j$}
            \STATE // Character-level representation
            \STATE $r^c = \text{CharCNN}(x_{\text{chars}})$
            \STATE // Word-level representation  
            \STATE $r^w = \text{WordEmbedding}(x_{\text{words}})$
            \STATE $h = \text{BiLSTM}([r^w; r^c])$
            \STATE // Emission scores
            \STATE $s = \text{Linear}(h)$
            \STATE // CRF loss
            \STATE $\mathcal{L} += -\log P(y|x; \theta)$
        \ENDFOR
        \STATE $\mathcal{L} = \mathcal{L} / |B_j|$
        \STATE $\theta = \theta - \eta \nabla_\theta \mathcal{L}$
        \STATE Clip gradients if $||\nabla_\theta \mathcal{L}|| > 5.0$
    \ENDFOR
    \STATE Evaluate on development set
    \STATE Apply early stopping if no improvement
\ENDFOR
\end{algorithmic}
\end{algorithm}

\subsection*{Code Availability}

The complete implementation is available at:
\url{https://github.com/TheAnig/NER-LSTM-CNN-Pytorch}

\end{document}